\documentclass{urac}
\usepackage[english,french]{babel}   
\usepackage[utf8]{inputenc}
\usepackage[pdftex]{graphicx}
\usepackage{hyperref}
\usepackage{tikz}
 
\usepackage[T1]{fontenc}

\usepackage{algorithm}
\usepackage{algpseudocode}

\titre{Étude de Problèmes d'Optimisation Combinatoire à Multiples Composantes Interdépendantes}

\auteur{\coord{Mohamed}{El Yafrani}{1}, \coord{Belaïd}{Ahiod}{1}}
\adresse{\affil{1}{LRIT, associated unit to CNRST (URAC 29)\\
Mohammed V University in Rabat}\\
B.P. 1014 Rabat, Morocco}
\email{m.elyafrani@gmail.com, ahiod@fsr.ac.ma}

\resumefrancais{
Ce résumé étendu présente une vue générale sur les problèmes d'optimisation NP-difficiles à 
multiples composantes interdépendantes. Ces problèmes se présentent dans plusieurs cas 
réels: industrie, ingénierie, et logistique. Le fait que ces problèmes sont souvent composés 
de deux ou plusieurs sous-problèmes NP-difficiles les rend plus difficiles à résoudre avec 
les méthodes exactes. Ceci est principalement à cause de la complexité élevée de ces 
algorithmes et la difficulté des problèmes eux-mêmes.

La source principale de difficulté de ces problèmes est la présence de dependances internes
entre les sous-problèmes. Cet aspet d'interdependance entre les composantes est présenté, 
et les grandes lignes pour des approches de résolution sont brievement introduites dans le contexte de 
(meta)heuristiques et calcul évolutionnaire.\\

\textbf{Mots-clés:} Interdépendance, Optimisation combinatoire, Métaheuristiques, Calcul évolutionnaire.
}

\resumeanglais{
This extended abstract presents an overview on NP-hard optimization problems with multiple 
interdependent components. These problems occur in many real-world applications: 
industrial applications, engineering, and logistics. The fact that these problems are 
composed of many sub-problems that are NP-hard makes them even more challenging to solve 
using exact algorithms. This is mainly due to the high complexity of this class of algorithms 
and the hardness of the problems themselves.

The main source of difficulty of these problems is the presence of internal dependencies between 
sub-problems. This aspect of interdependence of components is presented, and some outlines on 
solving approaches are briefly introduced from a (meta)heuristics and evolutionary computation 
perspective.\\

}

\newcommand\noteboxtext{%
  \footnotesize Extended abstract presented at the URAC days meeting in Rabat, Morocco.\\
  The meeting site is available at \url{https://sites.google.com/site/lriturac29/j-urac-2015}
}

\newcommand\noteboxnotice{%
\begin{tikzpicture}[remember picture,overlay]
\node[anchor=south,yshift=10pt] at (current page.south) {\fbox{\parbox{\dimexpr\textwidth\relax}{\noteboxtext}}};
\end{tikzpicture}%
}

\begin{document}

\maketitle

\noteboxnotice

\section{Motivation}
Les problèmes d'optimisation combinatoire interviennent dans plusieurs domaines. Parmi les 
domaines d'applications on retrouve l'ingénierie, les transports, et l'optimisation de la 
chaine d'approvisionnement. Ces problèmes sont souvent très difficiles à résoudre dans un 
temps raisonnable dû au fait qu'ils soient NP-difficiles. Néanmoins, ceci n'est pas le seul défi. 
Dans plusieurs cas réels le problème à 
optimiser est souvent composé de plusieurs sous-problèmes (dits composantes) qui sont eux-mêmes 
NP-difficiles. De plus, ces sous-problèmes sont interdépendants ce qui rend la tâche encore plus compliquée.

Ainsi, la difficulté dans les problèmes réels est due à plusieurs facteurs: NP-complétude, 
taille, et interdépendance. D'autres facteurs peuvent intervenir dans un problème réel 
\cite{michalewicz2013solve} tels que la dynamique de l'environnement, 
l'existence de contraintes, 
l'incertitude, pour ne citer que quelques-uns. Le but de ce travail est d’étudier ces facteurs, 
particulièrement l’interdépendance entres les composantes et la taille du problème, afin de 
concevoir des algorithmes efficaces de résolution.

\section{Problèmes à multiples\\ composantes interdépendantes}
\subsection{L'interdépendance entre composantes}
Un problème d'optimisation est dit à multiples composantes s'il se compose de deux ou plusieurs 
sous-problèmes. Une composante $B$ est dite dépendante d'une autre composante $A$ 
(On note $B \longleftarrow A$) si les deux condition suivantes sont satisfaites \cite{bonyadievolutionary}:
\begin{enumerate}
\item Générer une solution (non-forcement optimale) pour $A$ est possible sans l'aide 
de données produites par $B$.
\item Changer une solution de $A$ peut changer l'espace de recherche pour $B$. 
En d'autres termes, ceci peut changer la meilleure solution possible pour $B$ 
dans le contexte du problème général.
\end{enumerate}

\begin{figure}[ht]
\centering
\includegraphics[width=.45\linewidth]{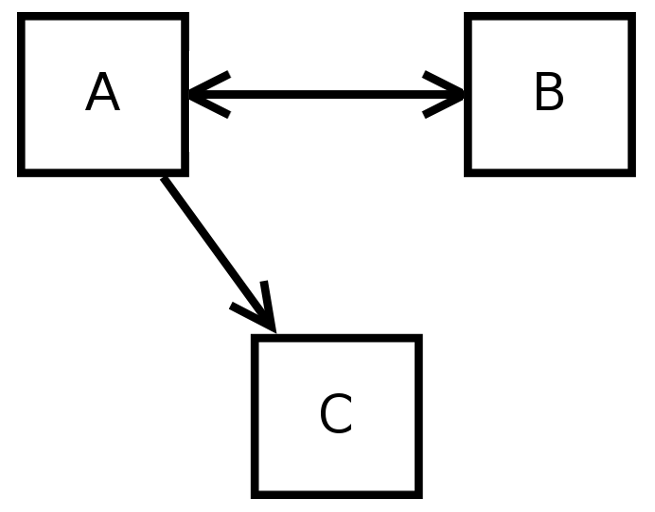}
\caption{Un problème à trois composantes \label{fig:3-problems}}
\end{figure}

Un problème à multiples composantes peut être représenté à l'aide d'un 
diagramme de dépendance. La figure \ref{fig:3-problems} illustre un problème à trois composantes interdépendantes. 
Dans cette figure, $C$ est dépendant de $A$, et $A$ et $B$ sont dépendants l'un de l'autre.

Dans un problème à plusieurs composantes interdépendantes, il ne suffit pas de résoudre les problèmes 
en isolation afin de résoudre le problème global. Une bonne stratégie de résolution doit prendre en 
considération l’interdépendance entre les différentes composantes du problème.

Parmis les problèmes à plusieurs composantes interdépendantes nous retrouvons:
le problème de l'assemblage du trafic, routage, et affectation de longueur d'onde \cite{hu2004traffic};
le problème du voyageur voleur \cite{ttp-2013, ttp-benchmark-2014};
le problème de localisation-routage \cite{perl1985warehouse};
et le problème de routage de véhicules sous contraintes de chargement \cite{iori2010routing}.

\subsection{Un exemple illustratif: Killer Sudoku}
\begin{figure}[ht]
\centering
\includegraphics[width=.96\linewidth]{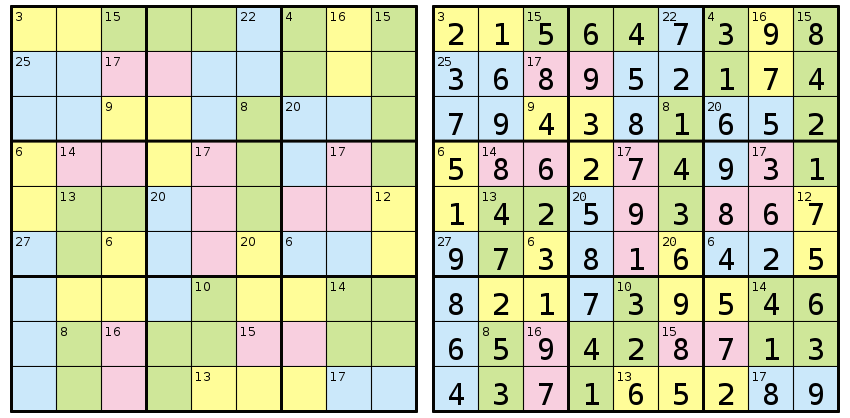}
\caption{
Un exemple de killer sudoku. 
Original images by Toon Spin (Toon81) (Own work) [Public domain], via Wikimedia Commons. 
Available at \url{https://en.wikipedia.org/wiki/Killer\_sudoku\#/media/File:Killersudoku\_color.svg} 
\label{fig:killer-sudoku}
}
\end{figure}

Le Killer Sudoku est un jeu de grille qui combine les deux jeux sudoku et kakuro 
\cite{sudoku-npc-2003}. 
Le but est de remplir la grille en respectant les règles du sudoku mais aussi celles du kakuro. 
La figure \ref{fig:killer-sudoku} illustre une instance de killer sudoku, ainsi qu'une solution possible.

L'interdépendance dans cet exemple est claire (dépendance dans les deux sens). 
Il est aussi clair que la résolution du sudoku en isolation ne résout pas le killer sudoku. 
De même, la résolution du kakuro sans prendre en compte les règles du sudoku ne résout pas 
le problème global.

\section{Solutions proposées}
\subsection{Jointure des voisinages}

Une première approche de résolution utilise la notion de voisinage dans un problème d'optimisation. 
L'approche consiste à appliquer un produit cartésien entre les voisinages des composantes afin de 
créer un voisinage propre au problème global. Le voisinage résultant pourrait être utilisé dans le 
contexte d'une recherche locale ou de métaheuristiques telles que 
la recherche locale itérative \cite{lourencco2003iterated} et le recuit simulé \cite{sa-83}.
L'algorithme \ref{algo:joint} illustre une 
telle approche pour le problème de la figure \ref{fig:3-problems}.

\begin{algorithm}[ht]
  \caption{ Un exemple d'un simple algorithme de recherche locale pour le problème de la figure \ref{fig:3-problems} } \label{algo:joint}
  \begin{algorithmic}[1]
    \State $(a,b,c) \gets solution initiale$
    \While{condition d'arrêt}
    
      \For{$a'$ voisin de $a$ dans $A$}
        \For{$b'$ voisin de $b$ dans $B$}
          \For{$c'$ voisin de $c$ dans $C$}
            \State évaluer $(a',b',c')$
            \If{$(a',b',c')$ est meilleure que $(a,b,c)$}
              \State $(a,b,c) \gets (a',b',c')$
            \EndIf
          \EndFor
        \EndFor
      \EndFor
    
    \EndWhile
  \end{algorithmic}

\end{algorithm}

Le problème de cette approche est la complexité de l'algorithme, ce qui induit un temps de calcul 
énorme pour les instances à grande taille.

\subsection{Cosolver}

Cosolver est un framework proposé exclusivement pour les problèmes à deux composantes 
interdépendantes \cite{socially-ttp-2014}. L’idée est d'isoler le traitement de chaque composante et de 
gérer une communication entre les solveurs: lorsque le solveur d'une composante $A$ trouve 
une meilleure solution, il la transmet au solveur de la composante $B$. De cette manière, 
l’interdépendance est gérée dans un processus itératif entre les deux "sous-solveurs". 
Ce processus est illustré dans le pseudo-code \ref{algo:cosolver}.

\begin{algorithm}[ht]
  \caption{Une implémentation simplifiée de Cosolver} \label{algo:cosolver}
  
  \begin{algorithmic}[1]
    \State $(a,b) \gets solution initiale$
    \While{condition d'arrêt}

      \State $a' \gets solveurA (a,b)$
      \State $b' \gets solveurB (a',b)$
      \State $(a,b) \gets (a',b')$

    \EndWhile
  \end{algorithmic}

\end{algorithm}

Cette approche est de nature parallélisable. En effet, le fait que les sous-solveurs 
actent de manière isolée permet de facilement de distribué le traitement.

\subsection{Algorithmes évolutionnaires}
Les algorithmes évolutionnaires \cite{john1992adaptation} 
sont bien adaptés pour attaquer les problèmes à multiples composantes interdépendantes. 
Ces méthodes offrent une excellente flexibilité, ils peuvent être utilisés de plusieurs 
manières. On pourrait concevoir un algorithme purement évolutionnaire pour 
résoudre le problème général, ou bien hybrider avec d'autres heuristiques. Par exemple, 
des opérateurs génétiques peuvent être inclus dans Cosolver afin d’améliorer sa performance 
en terme d'exploration de l'espace de recherche.

\section{Conclusion et perspectives}
Les problèmes réels sont bien plus difficiles que les problèmes académiques. Pourtant, 
la plupart des articles continuent de traiter des problèmes de benchmark 
(problème du sac-a-dos, bin packing, problème du voyageur de commerce, etc.) 
qui n'interviennent que sous forme de composantes d'un problème global. Peu 
de travaux se concentrent sur les spécificité de ces problèmes réels \cite{michalewicz2013solve}. 
Dans une perspective à court terme, nous nous engageons à concevoir des solutions 
heuristiques pour des cas particuliers de problèmes mono-objectifs à deux composantes. 
Nous voulons également continuer dans cette voie et généraliser nos approches pour 
des problèmes à plusieurs composantes. Nous avons prévu d'autres pistes de recherches 
prometteuses pour traiter ce type de problèmes telles que le parallélisme, le calcul 
distribué, et l'hybridation.

\small
\bibliographystyle{plain}
\bibliography{references}

\begin{thebibliography}{10}

\bibitem{ttp-2013}
Mohammad~Reza Bonyadi, Zbigniew Michalewicz, and Luigi Barone.
\newblock The travelling thief problem: the first step in the transition from
  theoretical problems to realistic problems.
\newblock In {\em Evolutionary Computation (CEC), 2013 IEEE Congress on}, pages
  1037--1044. IEEE, 2013.

\bibitem{bonyadievolutionary}
Mohammad~Reza Bonyadi, Zbigniew Michalewicz, Frank Neumann, and Markus Wagner.
\newblock Evolutionary computation for multi-component problems: Opportunities
  and future directions.
\newblock 2014.

\bibitem{socially-ttp-2014}
Mohammad~Reza Bonyadi, Zbigniew Michalewicz, Micha{\u{o}}
  Roman~Przyby{\u{o}}ek, and Adam Wierzbicki.
\newblock Socially inspired algorithms for the travelling thief problem.
\newblock In {\em Proceedings of the 2014 conference on Genetic and
  evolutionary computation}, pages 421--428. ACM, 2014.

\bibitem{john1992adaptation}
John~Henry Holland.
\newblock {\em Adaptation in natural and artificial systems: an introductory
  analysis with applications to biology, control, and artificial intelligence}.
\newblock MIT press, 1992.

\bibitem{hu2004traffic}
Jian-Qiang Hu and Brett Leida.
\newblock Traffic grooming, routing, and wavelength assignment in optical wdm
  mesh networks.
\newblock In {\em INFOCOM 2004. Twenty-third AnnualJoint Conference of the IEEE
  Computer and Communications Societies}, volume~1. IEEE, 2004.

\bibitem{iori2010routing}
Manuel Iori and Silvano Martello.
\newblock Routing problems with loading constraints.
\newblock {\em Top}, 18(1):4--27, 2010.

\bibitem{sa-83}
Scott Kirkpatrick, C~Daniel Gelatt, Mario~P Vecchi, et~al.
\newblock Optimization by simmulated annealing.
\newblock {\em science}, 220(4598):671--680, 1983.

\bibitem{lourencco2003iterated}
Helena~R Louren{\c{c}}o, Olivier~C Martin, and Thomas St{\"u}tzle.
\newblock {\em Iterated local search}.
\newblock Springer, 2003.

\bibitem{michalewicz2013solve}
Zbigniew Michalewicz and David~B Fogel.
\newblock {\em How to solve it: modern heuristics}.
\newblock Springer Science \& Business Media, 2013.

\bibitem{perl1985warehouse}
Jossef Perl and Mark~S Daskin.
\newblock A warehouse location-routing problem.
\newblock {\em Transportation Research Part B: Methodological}, 19(5):381--396,
  1985.

\bibitem{ttp-benchmark-2014}
Sergey Polyakovskiy, Mohammad Reza, Markus Wagner, Zbigniew Michalewicz, and
  Frank Neumann.
\newblock A comprehensive benchmark set and heuristics for the traveling thief
  problem.
\newblock In {\em Proceedings of the Genetic and Evolutionary Computation
  Conference (GECCO), Vancouver, Canada}, 2014.

\bibitem{sudoku-npc-2003}
YATO Takayuki and Seta Takahiro.
\newblock Complexity and completeness of finding another solution and its
  application to puzzles.
\newblock {\em IEICE transactions on fundamentals of electronics,
  communications and computer sciences}, 86(5):1052--1060, 2003.

\end{thebibliography}

\end{document}